\documentclass[sigconf]{acmart}

\AtBeginDocument{%
  }

\renewcommand\footnotetextcopyrightpermission[1]{}
\AtBeginDocument{%
  \addtolength{\footskip}{0pt}%
  \fancyhead{} 
  \fancyfoot{} 
  \pagestyle{empty} 
}

\usepackage{array}
\usepackage[most]{tcolorbox}
\usepackage{booktabs}      
\usepackage{tabularx}      
\usepackage{stfloats}      
\usepackage{float}         

\tcbset{
  companysample/.style={
    colback=gray!3,      
    colframe=black!15,   
    boxrule=0.3pt,       
    arc=1.5mm,           
    left=6pt,right=6pt,top=4pt,bottom=4pt,
    fonttitle=\bfseries\small,
    coltitle=black,
    before skip=6pt, after skip=6pt
  }
}

\begin{document}

\title{InterCorpRel-LLM: Enhancing Financial Relational Understanding with Graph–Language Models}

\author{Qianyou Sun}
\email{qysun@stu.pku.edu.cn}
\affiliation{%
  \institution{Guanghua School of Management, Peking University}
  \country{China}
}

\author{Jiexin Zheng}
\email{jiexinzheng@gsm.pku.edu.cn}
\affiliation{%
  \institution{Guanghua School of Management, Peking University}
  \country{China}
}

\author{Bohan Jin}
\email{bohanjin@stu.pku.edu.cn}
\affiliation{%
  \institution{Guanghua School of Management, Peking University}
  \country{China}
}

\author{Lihua Chen}
\email{chenlh@gsm.pku.edu.cn}
\affiliation{%
  \institution{Guanghua School of Management, Peking University}
  \country{China}
}

\author{Yijie Peng}
\email{pengyijie@gsm.pku.edu.cn}
\affiliation{%
  \institution{Guanghua School of Management, Peking University}
  \country{China}
}


\begin{abstract}
Identifying inter-firm relationships such as supply and competitive ties is critical for financial analysis and corporate governance, yet remains challenging due to the scale, sparsity, and contextual dependence of corporate data. Graph-based methods capture structure but miss semantic depth, while large language models (LLMs) excel at text but remain limited in their ability to represent relational dependencies. To address this, we propose InterCorpRel-LLM, a cross-modal framework that integrates GNNs with LLMs, supported by a proprietary dataset derived from FactSet supply chain records and three tailored training tasks: company graph matching, industry classification, and supply relation prediction. This design enables effective joint modeling of structure and semantics. Experiments show that InterCorpRel-LLM substantially outperforms strong baselines, including GPT-5, on a supply relation identification task, achieving an F-score of 0.8543 vs. 0.2287 with only a 7B-parameter backbone and lightweight training. The model also generalizes to zero-shot competitor identification, underscoring its ability to capture nuanced inter-firm dynamics. Our framework thus provides analysts and strategists with a robust tool for mapping and reasoning about complex corporate networks, enhancing decision-making and risk management in dynamic markets.
\end{abstract}


\maketitle

\section{Introduction}
\label{intro}

Understanding inter-firm relationships—particularly supply dependencies and competitive dynamics—is central to modern finance and corporate strategy, as these ties shape resilience, performance, and systemic risk \citep{wei2010strategic,ivanov2025digital,cao2025can}. Supply relations reveal the flow of goods and interdependencies across industries, where disruptions can cascade through networks and trigger financial shocks \citep{chen2022short,ivanov2025digital}. Competitive linkages, meanwhile, influence strategic positioning and innovation diffusion; failing to monitor them risks “competitive blind spots” that leave firms exposed to unforeseen threats \citep{pant2015web}. Accurately identifying such ties is therefore essential for financial analysis, risk management, and policy-making in an increasingly networked economy.

Modeling inter-firm relationships is challenging because the data are large-scale, sparse, and semantically rich. A single firm may connect to thousands of partners, yet networks are incomplete and context-dependent, with critical clues often buried in financial text or reports. Graph neural networks (GNNs) capture structural dependencies well but struggle when graphs are sparse or when meaning is encoded in language \citep{wasi2024graph,kosasih2022machine}. Conversely, large language models (LLMs) excel at interpreting nuanced text \citep{cao2025can} but lack mechanisms to reason over graph structure. Thus, graph-only methods miss context, while text-only methods miss structure, leaving each insufficient on its own.

Recent work has explored GNN-LLM hybrids that integrate structure and semantics, but existing designs have clear limitations. LLM-centric approaches linearize graphs and lose topology, while GNN-centric approaches compress text and sacrifice nuance \citep{yang2024gl}. Moreover, none have been adapted to the financial domain, where accurate modeling of supply chains and competition is essential. The absence of domain-specific frameworks and benchmarks has left analysts without effective tools, underscoring the need for tailored graph–language models for inter-firm relationship analysis \citep{wasi2024graph}.

To address this gap, we propose InterCorpRel-LLM, a GNN-LLM framework for modeling inter-firm relationships. Our method bridges structural and semantic modeling by combining graph data with domain-specific text. Using the FactSet Revere dataset, we construct a realistic domain-specific benchmark and design three complementary tasks: (1) Company Graph Matching to ground language in network context, (2) Industry Classification to capture high-level semantic groupings, and (3) Supply Relation Prediction to learn inter-company business relationships. Together, these tasks with lightweight training address core challenges in entity resolution, semantic understanding, and inter-firm relationship identification and enable the model to embed firms in a shared structural–textual space, capturing nuanced business relationships more effectively.

Our experiments show that InterCorpRel-LLM achieves substantial gains in both supply relation prediction and competitor identification, even without modifying the parameters of the backbone model. Instead of fine-tuning the large-scale language model itself, we train only a lightweight graph–language projector that aligns structural and textual representations. This design not only ensures parameter efficiency but also demonstrates that meaningful structural–semantic integration can be achieved without costly full-model optimization. Despite its compact 7B-parameter backbone, InterCorpRel-LLM surpasses much larger open- and closed-source models—including GPT-5 and DeepSeek-v3.1—achieving an F-score of 0.8543 versus 0.2287 in the supply-relation task. In the competitor identification task, our dedicated training scheme further enhances relational understanding, enabling robust zero-shot rival detection. These results confirm that \textbf{carefully designed graph–language fusion with domain-specific supervision}, rather than brute-force scaling, is the key to efficient and accurate inter-firm relationship modeling.

The main contributions of this work are summarized as follows:
\begin{itemize}
    \item We propose \textit{InterCorpRel-LLM},  domain-specific GNN–LLM architecture that unifies graph topology and financial text for inter-firm relationship modeling. Despite using only a 7B-parameter backbone, our model achieves state-of-the-art performance, surpassing GPT-5 and DeepSeek-v3.1 in key task such as supply relation prediction.
    \item We construct a supply-chain-based benchmark with three complementary tasks—graph matching, industry classification, and supply relation prediction—covering structural grounding, semantic grouping, and relational inference.
    \item We develop a parameter-efficient multi-task fine-tuning scheme that jointly optimizes structural and textual representations, achieving strong generalization in firm-relation understanding task without large-scale pretraining.
\end{itemize}

\section{Related Works}

\textbf{Supply Chain Relation Identification.}
\citet{kosasih2022machine} addressed supply chain opacity with a graph neural network (GNN)–based approach, relying solely on structural information while neglecting firm-specific context. Although effective on an automotive supplier dataset, their method has not been validated across broader industries. An alternative line of work leverages corporate reports or other public data to infer missing supply chain links. For example, \cite{wichmann2020extracting} applied natural language processing (NLP) techniques to online data, but their reliance on traditional models such as BiLSTM and SVM limited generalization and performance, particularly outside the automotive and aerospace sectors. More recently, \cite{jin2025enhancing} introduced a paradigm that harnesses large language models (LLMs) to extract supply chain relations via contextual question–answering. This approach significantly improves generalization across industries, though it remains ineffective when firms deliberately conceal relationships absent from public sources.

\textbf{Competitor Identification.}
Accurate recognition of rivals is central to valuation, governance, and strategic forecasting. Traditional approaches—such as industry codes \citep{phillips2016industry}, managerial judgment, text-based similarity \citep{hoberg2010product,hoberg2016text}, or curated databases—often suffer from incompleteness and “competitive blind spots” \citep{pant2015web}. To overcome these limitations, \cite{pant2015web} introduced online isomorphism, showing that overlapping web content and hyperlink structures can effectively signal competition, outperforming offline baselines. More recently, \cite{cao2025can} demonstrated that large language models (LLMs) capture nuanced inter-firm similarities beyond conventional text-mining. Despite these advances, existing methods remain constrained in scalability, cross-industry generalization, and integration with structured financial networks. Developing approaches that jointly exploit network structure and semantic reasoning thus remains an open challenge for automated competitor identification.

\textbf{GNN–LLM Integration.}Recent studies have explored GNN–LLM integration in various domains. \cite{wang2025exploring} showed that pure LLMs underperform on graph tasks without explicit structure, motivating hybrid approaches. In recommendation field, \cite{xi2024towards} proposed KAR to augment models with reasoning and factual knowledge from LLMs, while \cite{wei2024llmrec} combined graph signals with LLM-based user–item representations. For open-ended graph reasoning, \cite{tang2024graphgpt} introduced GraphGPT with instruction tuning to align LLMs to graph structures, and \cite{zhang2024graphtranslator} developed GraphTranslator to bridge pretrained GNNs with LLMs for both predefined and open-ended tasks, such as paper citation prediction and product recommendation. These advances highlight the promise of combining structural and semantic modeling, yet remain limited to several fields mentioned above. To our knowledge, we are the first to adapt a GNN–LLM framework to inter-company relation modeling, addressing the unique challenges of supply chain and competitor analysis in finance.

\section{Approach}
\label{approach}

\subsection{Preliminaries}
In this study, the task of identifying inter-firm business relationships is formulated 
based on the supply chain network of firms. Specifically, the inference of supply and 
competitive relations between firms is conducted with the known supply chain network 
as the foundation. The supply chain network of firms can be viewed as a special type 
of graph data. Compared to citation networks or product recommendation networks, 
supply chain data is enriched with more complex contextual information, potentially
involving corporate operations, geopolitical conditions, and broader socioeconomic
environments. This intrinSIC complexity fundamentally determines the difficulty of 
inferring inter-firm business relationships.  

The identification of inter-firm supply relations can be regarded as a \textbf{directed edge 
prediction problem} in the supply chain network. The network is represented as
\begin{equation} \label{eq:supply_chain_graph}
G = (V, E, X),
\end{equation}
where $V$ denotes the set of nodes, with each node representing a firm. The cardinality 
$|V| = N$ indicates the total number of firms. $E$ denotes the set of directed 
edges in the network; if there exists an edge from node $k$ to node $g$, this represents 
a supply relation where firm $k$ provides certain products or services to firm $g$. 
The node feature matrix is denoted as $X \in \mathbb{R}^{N \times F}$, where each row corresponds to a 
firm and $F$ is the dimension of the node feature vector. The supply relation prediction 
task is then defined as predicting the missing edges in $E$, given $V$, $X$, and 
the known subset of $E$.  

To test whether our model truly understands the complex business relationships between enterprises, 
we also designed the task of identifying company competitors.
The identification of inter-firm competitive relations can be formulated as a 
\textbf{binary classification task} grounded in information from the supply chain 
network. In this setting, all components of the supply chain graph 
$G = (V, E, X)$ 
are assumed to be known, and the goal is to predict 
whether a pair of firms forms a competitive relationship.

\subsection{Data Encompassing Comprehensive Information on Corporate Relationships}

To facilitate effective learning of inter-firm business relationships, we construct a domain-specific dataset from FactSet’s Supply Chain Relationships records. We extract a directed graph of U.S. public companies from 2023, where nodes are firms and edges represent a known supplier→customer link (as documented by FactSet). The resulting supply network contains \textbf{3,211 firms spanning a wide range of sectors} and 11,635 verified inter-firm supply links. This graph provides the structural backbone of our data. FactSet’s supply chain data is known to capture multi-tier supplier/customer networks and is used to uncover hidden dependency risks, making it an ideal foundation for our task. 

We augment each company node with rich textual and categorical information to provide context that pure graph structure alone would miss. In particular, for each firm we include:

\textbf{Business Description (Annual Report Text)}: Each firm’s 2023 annual report (10-K filing) is processed to extract a synopsis of its core business activities, products, and financial highlights. These unstructured texts typically detail what the company does, its revenue streams, and its scale of operations. Such information is essential for understanding a firm’s strategic positioning and the nature of its relationships. 

\textbf{Geographic Location}: We attach the firm’s headquarter location (country/region). Geographic context can be important, as supply chain ties often have regional patterns and risks (e.g. proximity can matter for certain logistics, and regulations/trade policies differ by region).

\textbf{Industry Classification}: Each firm is labeled with its industry category under the Standard Industrial Classification (SIC) system. SIC codes provide a hierarchical industry grouping (e.g. a firm might be categorized as SIC 3674: Semiconductors and Related Devices). This gives the model a structured indication of the company’s sector. Industry labels are valuable for relationship reasoning because, for instance, supply links typically connect firms in related industries (automotive manufacturers will link to auto parts suppliers, etc.), and competitors are usually in the same industry. 

By combining the structural graph of who-supplies-whom with each firm’s textual and categorical profile, our dataset provides a holistic view of the corporate ecosystem. This richness enables training tasks that require understanding not just network connections but also the business context behind those connections.

For competitor identification task, we obtain a competitor dataset from a commercial data provider that covers all firms in the above supply chain network and their pairwise competitive relations. This dataset enables us to evaluate whether the model can effectively identify competitors in a zero-shot setting, thereby testing its capacity to capture broader inter-firm dynamics.

\subsection{InterCorpRel-LLM: Graph–Language Model for Company Relation Identification}

\captionsetup[figure]{skip=\baselineskip, belowskip=\baselineskip}

\begin{figure*}[ht] 
  \centering
  \includegraphics[width=\linewidth]{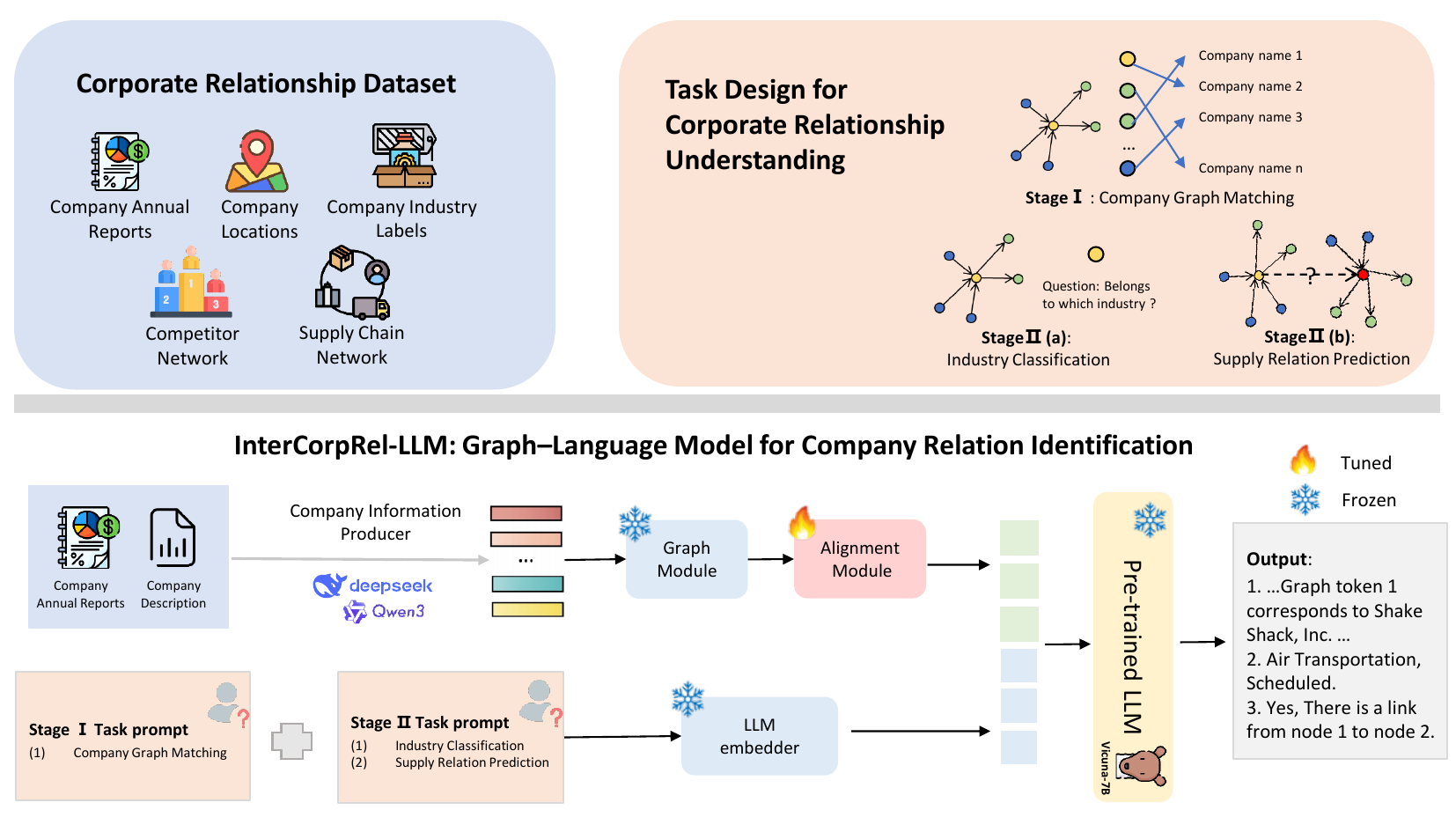}
  \caption{The overall structure of InterCorpRel-LLM}
  \label{fig:method-pipeline}
\end{figure*}

Inspired by prior research\citep{tang2024graphgpt,zhang2024graphtranslator}, the proposed framework consists of the following key components( \autoref{fig:method-pipeline}):

\begin{itemize}
\item(1) a \textit{Company Information Producer}, which extracts and summarizes firm-level background information into vector representations; 

\item(2) a \textit{Graph Module}, which captures structural information from the supply chain network based on the vector representations from \textit{Company Information Producer}; 

\item(3) an \textit{Alignment Module}, which maps graph-based embeddings into a representation space (graph tokens) more compatible with large language models (LLMs); 

\item(4) an \textit{LLM Module}, which integrates graph-structured (graph tokens as part of the task prompt) and unstructured textual modalities for inter-company relation identification. For details on the prompt design, please refer to the attachment.
\end{itemize}

\textbf{Company Information Producer.} Leveraging LLMs to enhance data representations has been widely recognized as an effective approach for improving model performance\cite{hong2025llm}. In this study, we design a Company Information Producer that generates enriched node representations for each firm in the network. First, we employ the proprietary model DeepSeek-V3\citep{liu2024deepseek} to summarize annual reports of publicly listed firms (details provided in the Appendix). Subsequently, we utilize the Qwen3-Embedding-8B model to produce 128-dimensional vector representations of firm nodes ($F=128$). The Qwen3 series incorporates Multi-resolution Learning (MRL) during training, which ensures that truncating to the first $n$ dimensions does not result in linear semantic loss; rather, the leading dimensions retain condensed semantic information\citep{zhang2025qwen3}.  

\begin{tcolorbox}[companysample, title={Company Introduction Sample}]
\small
\textbf{Company Name:} ConnectOne Bancorp, Inc.\\
\textbf{Introduction:} ConnectOne Bancorp, Inc. is a financial holding company that operates primarily through its subsidiary, ConnectOne Bank, a New Jersey-chartered commercial bank. The company provides a comprehensive range of banking services, including deposit and loan products, to small and mid-sized businesses, professionals, and individuals in the New York Metropolitan area and South Florida. 

ConnectOne Bank offers personal and business checking accounts, savings and money market accounts, time deposits, and retirement accounts. Its lending portfolio includes commercial loans, commercial real estate loans, residential mortgages, home equity loans, and consumer loans. The bank also provides cash management services, digital banking solutions, and merchant services. 

Through its fintech subsidiary, BoeFly, ConnectOne facilitates business lending by connecting borrowers in the franchise sector with a network of partner banks. The company operates a branch-lite model, leveraging technology and regional offices to serve clients efficiently. ConnectOne Bancorp focuses on relationship-driven banking with a strong emphasis on personalized service and competitive financial solutions.
\end{tcolorbox}

\textbf{Graph Module.} This module adopts a contrastive learning framework inspired by the CLIP (Contrastive Language–Image Pretraining) architecture, but applied to graph and text modalities. The objective is to align graph-based and text-based representations of the same firm while separating those of different firms. This encourages the firm embeddings to not only capture supply chain structural information but also reside closer to the textual embedding space. In this study, we employ the pretrained model released by \cite{tang2024graphgpt}, who demonstrated that their model, trained on citation networks and product purchase data, effectively transmits graph information into embeddings that can be further processed by LLMs.

\textbf{Alignment Module.} To ensure that graph embeddings generated by the GNN module are effectively embedded into the LLM, we introduce an alignment layer between the Graph Model and the LLM Module, which maps the graph representations into a space more compatible with LLM inputs.  Alignment Module is one fully connected layer.

\textbf{LLM Module.} Finally, the LLM integrates the graph-modality information from the Alignment Module with unstructured textual descriptions of firms to predict inter-firm relations. In this study, we employ \texttt{Vicuna-7B-v1.5} as the backbone LLM, consistent with numerous prior works on graph–language model integration\citep{cao2023instructmol,wang2024llms,wu2024exploring}.

\subsection{Task Design for Corporate Relationship Understanding}

We devise three focused training tasks, organized into a two-stage fine-tuning paradigm, to inject the graph-structured supply chain information into the LLM’s representations. The goal is to enhance the model’s semantic understanding of inter-firm relationships without altering the pre-trained GNN or LLM weights. Only a small Alignment Module (which maps graph embeddings to the LLM input space) is updated during training. This design keeps the majority of the model’s knowledge intact while infusing new relational reasoning abilities.

\textbf{Stage I: Company Graph Matching.}  
The objective of this stage is to enable the LLM to better interpret 
supply chain structural signals processed by the Graph Module (GM) 
and Alignment Module. By training a lightweight cross-modal mapping, 
the model reduces its reliance on labeled data in later supervised tasks 
and achieves faster convergence. Specifically, we extract subgraphs centered on a target company with its 1-hop neighborhood and 
obtain vectorized representations using the GM. 
These are mapped via the Alignment Module into a sequence of graph tokens, 
with the number of tokens equal to the subgraph’s node count. The task is 
formulated as a graph–text matching problem: aligning graph tokens with 
their corresponding firm textual identifiers (standardized company names). 
Formally, the probability of generating output $O$ is:  
\begin{equation} \label{eq:prob_O}
P(O \mid S_H, S_T) = \prod_{i=1}^{L} p_{\theta}\big(x_i \mid S_H, S_{T,<i}, O_{<i}\big),
\end{equation}
where $S_H$ denotes the graph token sequence processed by the GM and 
Alignment modules, $S_T$ is the company name sequence, $L$ is the text 
length, and $\theta$ represents learnable parameters within the framework. 
Importantly, this stage is self-supervised (the “ground truth” name comes from the data itself), 
which helps the model bootstrap an alignment without needing labeled examples of supply links. 
After Stage I, the model has learned a preliminary joint space for graph and text:
the graph-structured signals about a company and its neighbors start to influence the LLM’s internal representations.

\textbf{Stage II: Industry Classification and Supply Relation Prediction.}  

In the second stage, we jointly fine-tune the model on two supervised tasks that are highly 
relevant to business understanding: predicting a company’s industry sector, 
and predicting supply chain links between companies. 
These tasks further refine the model’s grasp of industry knowledge and relational logic.

\textbf{Industry Classification Task:} Each training instance consists of a single firm’s subgraph 
(the firm plus its immediate neighbors in the supply chain network, as before) 
and the textual description of that firm (from its profile or annual report). 
The model is asked to output the firm’s industry category (a specific SIC sector name). 
This is essentially a text classification task aided by graph context. 
The presence of certain neighbors in the graph can hint at the industry (for instance, if a company’s neighbors include car parts suppliers and dealerships, the company is likely in the automotive sector). 
By learning this task, the model internalizes which features (both in text and graph) correlate with industry labels. 
This encourages the LLM to incorporate structural cues into its understanding of the firm’s business domain.

\textbf{Supply Relation Prediction Task:} Here the model is given a pair of firms $(A, B)$ 
along with their respective 1-hop neighbor subgraphs (excluding each other). 
The textual inputs include both firms’ names and descriptions, 
plus the graph tokens from their subgraphs (via the Alignment Module). 
The task is to predict whether a directed supplier→customer relation exists from $A$ to $B$. 
We formulate this as a binary classification or a conditional text generation (e.g. generating “Yes” if $A$ supplies $B$, otherwise “No”). 
Successful prediction requires the model to analyze the compatibility of the two firms’ profiles and their network contexts. For example, if $A$ makes auto engines and $B$ is an automobile manufacturer, and $A$ is not already a neighbor of $B$, the model might predict a supply relation is plausible; 
conversely, if $A$ is a retail store and $B$ is a mining company with no common industry ground, it should predict no supply relation. 
Through this task, the model learns the structural logic of supply chains – which industry pairs typically form supplier-customer links, what patterns of shared neighbors or attributes indicate a likely connection, etc. 

During Stage II, the model is optimized on both tasks. Notably, we keep the GNN and the core LLM frozen; only the Alignment Module is trainable. This strategy ensures we \textbf{preserve the pretrained knowledge} in the LLM (e.g. general language and commonsense, plus any financial knowledge it already has\citep{you2025large}) and in the GNN which could be pre-trained on generic graph structures, while just \textbf{infusing new domain-specific abilities}. It also avoids overfitting given our limited labeled data – by leveraging self-supervision (Stage I) and light supervised signals (Stage II), we achieve fast convergence with minimal data.

After fine-tuning, the unified model is expected to not only recall factual supply links, but also reason about them: it should infer likely relationships even between companies it has never seen before, based on their profiles and network context, and understand broader concepts like industry competition.

\section{Result}

\subsection{experiment setup}

\begin{table*}[t]
\centering
\footnotesize
\setlength{\tabcolsep}{3pt}
\renewcommand{\arraystretch}{1.2}
\caption{Prompt for InterCorpRel-LLM tasks}
\label{tab:prompt-intermodel}
\begin{tabular*}{0.97\textwidth}{@{\extracolsep{\fill}}p{0.18\textwidth}p{0.77\textwidth}@{}}
\toprule
\textbf{Task} & \textbf{Prompt Template} \\
\midrule
\textbf{Company Graph Matching} &
\textit{Instruction:} Given a sequence of graph tokens \textless graph\textgreater{} that constitute a subgraph of an industry supply chain graph, where the first token represents the central node of the subgraph, and the remaining nodes represent the first order neighbors of the central node. Each graph token contains the content of the introduction of the company. If a company supplies some product or service to another one, a link is constructed from the former to the latter. Here is a name list of companies: \{company\_names\}, please reorder the list according to the order of graph tokens (i.e., complete the matching of graph tokens and company names).

\textit{(\textless graph\textgreater{} is a placeholder that will be replaced by graph tokens generated by the Alignment Module. \{company\_names\} is the text sequence of company names to be matched.)}
\\[4pt]

\textbf{Industry Classification} &
\textit{Instruction:} Given an industry supply chain graph: \textless graph\textgreater where the 0th node is the central company, and other nodes are its one-hop or multi-hop neighbors, with the following information: Description: \{ company\_description \} Company name: \{ company\_name \} Question: Which industry category does the company belong to under the SIC classification system? Give the most likely SIC industry category of this company directly, in the form "full name of the category" .

\textit{(\textless graph\textgreater  is a placeholder that will be replaced by graph tokens generated by the Alignment Module. \{company\_description\} is the focal company textual description. \{company\_name\} is the the focal company's name.)}
\\[4pt]

\textbf{Supply Relation Prediction} &
\textit{Instruction:} Given a sequence of graph tokens: \textless graph1\textgreater that constitute a subgraph of an industry supply chain graph, where the first token represents the central node of the subgraph, and the remaining nodes represent the first order neighbors of the central node. The information of the central node is as follow: Description: \{company1\_description\} Company name: \{company1\_name\}. and the other sequence of graph tokens: \textless graph2\textgreater , where the first token (the central node) with the following information: Description: \{company2\_description\}. Company name: \{company2\_name\}. If the link from node 1 to node 2 represent the supply chain relationship from the former company to the latter company, is there a link from node 1 to node 2? Give me a direct answer of "yes" or "no".

\textit{(\textless graph1/2\textgreater  is a placeholder that will be replaced by graph tokens generated by the Alignment Module. \{company1/2\_description\} is the focal company textual description. \{company1/2\_name\} is the the focal company's name.)}
\\[4pt]

\textbf{Competitor Identification} &
\textit{Instruction:} Given a sequence of graph tokens: \textless graph1\textgreater that constitute a subgraph of an industry supply chain graph, where the first token represents the central node of the subgraph, and the remaining nodes represent the first order neighbors of the central node. The information of the central node is as follow: Description: \{company1\_description\} Company name:  \{company1\_name\}. and the other sequence of graph tokens: \textless graph2\textgreater , where the first token (the central node) with the following information: Description:  \{company2\_description\}. Company name: \{company2\_name\} . If the link of the industry supply chain graph represents a supply chain relationship from the source company to the target company, whether the two companies represented by the central nodes of the two subgraphs are competitors of each other in business? Give me a direct answer of "yes" or "no".

\textit{(\textless graph1/2\textgreater  is a placeholder that will be replaced by graph tokens generated by the Alignment Module. \{company1/2\_description\} is the focal company textual description. \{company1/2\_name\} is the the focal company's name.)}
\\
\bottomrule
\end{tabular*}
\end{table*}

We evaluate our approach on two types of inter-firm relationship tasks: supply chain link prediction and competitor identification. To simulate a realistic scenario, we partition the supply chain graph data at the firm level with a $9{:}1$ train–test split, ensuring that all companies in the test set are entirely unseen during training . For the supply link prediction task, we further divide test links into two scenarios\citep{zhang2022link,baek2020learning,albooyeh2020out}:

(i) \textit{Inductive / semi-inductive link prediction}, At least one endpoint of the link was seen during training. The inductive test set contains 1,954 positive supplier-customer links and 1,954 randomly sampled negative pairs (firm pairs with no supply relation). 

(ii) \textit{Fully inductive link prediction}, both firms in the link are unseen during training. This fully inductive test set is much smaller, with 95 positive links and 95 negatives.  

This split allows us to assess generalization to completely new firms. In the fully inductive scenario, models cannot rely on any pre-existing supplier-customer link information for either node.

For the competitor relation prediction task, we use competition data 
recorded in a commercial dataset for the selected set of 3,211 firms in 
2023, comprising a total of 8,895 competitive relations. From these, we 
randomly sample 2,000 positive relations and 2,000 negatives (non-existent 
competitor pairs) to construct the dataset. This task is used purely to test the zero-shot transferability of inter-firm reasoning learned from the supply chain domain.

\subsection{Baseline Methods}

We compare our proposed InterCorpRel-LLM framework against three categories of baseline methods:  

(1) \textbf{Traditional GNN baselines:} We train graph neural networks on the observed supply chain graph structure. In particular, we use a Graph Attention Network (GAT)\citep{velickovic2017graph}and a GraphSAGE model\citep{hamilton2017inductive}. Both are enhanced with initial node embeddings derived from \textbf{Company Information Producer} that encodes firm background features  into each node. This provides the GNNs some textual and attribute information to complement the graph topology.

(2) \textbf{Open-source LLMs:} We evaluate Vicuna-7B-v1.5-16k and Vicuna-13B-v1.5-16k, two open large language models, in a zero-shot manner. The prompt includes firm names, their business descriptions (from annual reports), geographic locations, and a structured context of their known suppliers/customers (formatted as a short text). We craft two prompt variants for each query: one without industry labels and one with industry labels (company’s SIC sector name) appended to see if explicit industry context helps.

(3) \textbf{Closed-source LLMs:} We also test powerful proprietary models – DeepSeek-V3.1, GPT-4o, and GPT-5 – using the same zero-shot prompt templates (with and without SIC industry labels). These models represent state-of-the-art black-box LLMs and serve as an upper bound for language-only reasoning on our tasks.

In addition, we include ablations of our own model to gauge the impact of each training stage. Our full \textbf{InterCorpRel-LLM} is trained in two stages: Stage I aligns company names with the supply graph (via graph-text matching), and Stage II fine-tunes the model on the relational tasks. We consider two variants:

\begin{itemize}
    \item \textbf{InterCorpRel-LLM\_CGM}: A partial training baseline that is only trained on Stage I (Company Graph Matching) and not fine-tuned on any Stage II relational tasks. 
    \item \textbf{InterCorpRel-LLM\_IC\_SRP}: Our model trained on Stage I, then Stage II with a multi-task objective combining Industry Classification(IC) and Supply Relation Prediction(SRP). This variant injects industry-related supervision during training in addition to the supply relation task.
\end{itemize}

\subsection{Performance in supply chain relationship identification}

\begin{table*}[t]
\centering
\footnotesize
\setlength{\tabcolsep}{3pt}
\renewcommand{\arraystretch}{1.2}
\caption{Company supply relation prediction result}
\label{tab:supply-rel}
\begin{tabular*}{0.95\linewidth}{@{\extracolsep{\fill}} p{4cm} *{8}{c} @{}}
\toprule
\multicolumn{1}{c}{} &
\multicolumn{4}{c}{\textbf{without SIC label}} &
\multicolumn{4}{c}{\textbf{with SIC label}} \\
\cmidrule(lr){2-5}\cmidrule(l){6-9}  
\multicolumn{1}{c}{\textbf{Model}} &
\multicolumn{2}{c}{Inductive} &
\multicolumn{2}{c}{Fully inductive} &
\multicolumn{2}{c}{Inductive} &
\multicolumn{2}{c}{Fully inductive} \\
\cmidrule(lr){2-3}\cmidrule(lr){4-5}\cmidrule(lr){6-7}\cmidrule(l){8-9}
 & Acc. & F-score & Acc. & F-score & Acc. & F-score & Acc. & F-score \\
\midrule
\multicolumn{9}{@{\hspace{39pt}}l}{\textit{GNN baselines}}\\
GAT   & 0.7526 & 0.5359 & 0.6613 & 0.1887 & -- & -- & -- & -- \\
SAGE  & 0.7774 & 0.5390 & 0.7253 & 0.0000 & -- & -- & -- & -- \\
\midrule
\multicolumn{9}{@{\hspace{36pt}}l}{\textit{LLMs and our models}}\\
vicuna-7b-v1.5  & 0.7377 & 0.7736 & 0.6579 & 0.5912 & 0.7326 & 0.7111 & 0.5895 & 0.4534 \\
vicuna-13b-v1.6 & 0.6400 & 0.3973 & 0.6421 & 0.4040 & 0.6379 & 0.3891 & 0.6000 & 0.3117 \\
deepseek-v3.1   & 0.5386 & 0.1375 & 0.5000 & 0.0000 & 0.5676 & 0.2216 & 0.5263 & 0.0957 \\
GPT4o           & 0.5827 & 0.2680 & 0.5421 & 0.1458 & 0.0957 & 0.2692 & 0.5263 & 0.5827 \\ 
GPT5            & 0.5548 & 0.1893 & 0.5263 & 0.0957 & 0.5660 & 0.2287 & 0.5316 & 0.1294 \\ 
\addlinespace  
\textit{InterCorpRel-LLM\_CGM}     & 0.3930 & 0.5639 & 0.4053 & 0.5725 & 0.3984 & 0.5658 & 0.4105 & 0.5692 \\
\textbf{InterCorpRel-LLM\_IC\_SRP} & \textbf{0.7968} & \textbf{0.8286} & \textbf{0.8105} & \textbf{0.8393} & \textbf{0.8347} & \textbf{0.8543} & \textbf{0.8368} & \textbf{0.8517} \\
\bottomrule
\end{tabular*}

\vspace{0.5ex}
\begin{minipage}{0.95\linewidth}\footnotesize
\emph{Notes.} "without/with SIC label" settings is only for LLMs and our models. "without SIC label" refers to predictions without industry code information, while "with SIC label" incorporates SIC labels. 

"Inductive" predicts links between a firm that was not observed during training and a firm that was included in the training data; "Fully inductive" predicts links between two firms that were both absent from the training phase. 
Following our evaluation protocol, F-score is the primary metric; accuracy is reported for completeness.
\end{minipage}
\end{table*}

Traditional graph neural networks (GraphSAGE and GAT) achieve moderately good accuracy on the inductive link prediction task  (around 75–78\% accuracy when at least one firm was seen in training). \textbf{However, their F1-scores are significantly lower, especially in the fully inductive scenario.}(\autoref{tab:supply-rel}) GAT’s F1 drops to only 0.19 and SAGE’s F1 drops to near 0 when predicting links between two completely unseen firms during training. This highlights the difficulty these structure-only models have in generalizing to entirely new nodes, even with company feature embeddings , a known weakness for GNNs in out-of-distribution settings. Overall, the GNN baselines can memorize and interpolate within the training graph, but they falter in the face of novel firms and links, as evidenced by their F1 on the fully inductive links task.

Zero-shot LLM baselines, despite their impressive general language understanding, underperform markedly on this structured link prediction task. Vicuna-7B and Vicuna-13B do show some capability on the inductive subset, for instance, Vicuna-7B achieves about 77\% F1 inductively (without SIC labels). Yet, both Vicuna models degrade on the fully inductive set (Vicuna-7B F1 $\approx$ 0.59, and Vicuna-13B only 0.40). Interestingly, the 7B model outperforms the 13B model in our tests; the larger Vicuna may be overfitting to irrelevant textual patterns in the prompt instead of truly understanding the graph context, whereas the 7B model, being simpler, might generalize slightly better. The closed-source LLMs (DeepSeek-V3.1 and the GPT-4o/GPT-5 class models) fare even worse – they nearly fail completely to identify true supply links (e.g. DeepSeek-V3.1 achieves near-zero F1 on fully inductive links). Moreover, including the SIC industry labels in the prompt had mixed effects: in some cases it provided a marginal benefit, but often it made little difference or even confused the LLMs. For example, Vicuna-7B’s inductive F1 decreased slightly from 0.7736 to 0.7111 when SIC codes were added. The inconsistent impact of adding industry label context suggests that these off-the-shelf LLMs are not adept at incorporating structured business metadata when simply given as extra text in a prompt.

\begin{table*}[!t]
\centering
\footnotesize
\setlength{\tabcolsep}{3pt}
\renewcommand{\arraystretch}{1.0}
\caption{Company competitor relation prediction result}
\label{tab:competitor-prediction}
\begin{tabular*}{0.85\linewidth}{@{\extracolsep{\fill}} p{4cm} *{4}{c} @{}}
\toprule
\multicolumn{1}{c}{\textbf{Model}} &
\multicolumn{2}{c}{\textbf{without SIC label}} &
\multicolumn{2}{c}{\textbf{with SIC label}} \\
\cmidrule(lr){2-3}\cmidrule(l){4-5}
 & Acc. & F-score & Acc. & F-score \\
\midrule
vicuna-7b-v1.5       & 0.7258 & 0.6347 & 0.7362 & 0.6535 \\
vicuna-13b-v1.5      & 0.5895 & 0.5252 & 0.5805 & 0.4994 \\
\addlinespace  
\textit{InterCorpRel-LLM\_CGM}      & 0.4150 & 0.5534 & 0.4150 & 0.5545 \\
\textbf{InterCorpRel-LLM\_IC\_SRP} & \textbf{0.7855} & \textbf{0.7713} & \textbf{0.7965} & \textbf{0.7774} \\
\bottomrule
\end{tabular*}

\vspace{0.5ex}
\begin{minipage}{0.85\linewidth}\footnotesize
\emph{Notes.} "without SIC label" refers to predictions without industry code information, 
while "with SIC label" incorporates SIC labels. 
Following our evaluation protocol, F-score is the primary metric; accuracy is reported for completeness.
\end{minipage}
\end{table*}

\textbf{In stark contrast, our fine-tuned models achieve dramatically higher performance. }
The fully-trained \\ InterCorpRel-LLM\_IC\_SRP attains an inductive F1 of 0.8543 and a fully-inductive F1 of 0.8517 with SIC label – the highest in each category by a large margin. And steady performance improvement can be found when the SIC information is added. \textbf{This indicates that our two-stage training effectively teaches the model how to reason about supplier relationships even for entirely new companies, while accurately leveraging industry SIC information when available and still maintaining strong performance in its absence.} Notably, the partially trained InterCorpRel-LLM\_CGM model (Stage I only) performs significantly worse than the full model – its F1 scores hover around 0.56–0.57 in all cases, barely better than random guessing. This confirms that while Stage I is a crucial foundation, it is insufficient by itself for the complex task of predicting supply links. The model needs the Stage II task-specific fine-tuning to truly learn the semantics of inter-firm relationships.

These results clearly demonstrate that neither structure-only methods nor text-only methods are adequate for accurate supply chain link prediction in a sparse, dynamic business network. The \textbf{InterCorpRel-LLM} outperforms all baselines by a wide margin,especially in correctly identifying true supply relationships, highlighting the value of our hybrid approach. By integrating structural graph context into a language model and training with domain-specific objectives, the model learns to leverage both data modalities. The superior F1 scores of \textbf{InterCorpRel-LLM}, even on entirely new firms, underscore the necessity of domain-adaptive alignment – effectively teaching the model the “language” of company supply networks – to achieve high fidelity in inter-firm relationship modeling.

\subsection{Performance in zero-shot competitor relationship identification}

To test our model's generalization ability in identification of other business inter-firm relationships, we test \textbf{InterCorpRel-LLM} on a zero-shot task of identifying whether two firms are competitors, without any fine-tuning. In this task, the input given to the model is the same as that in the supply relation prediction task, including customized structure information and company text information of the two focal companies. However, in this task, the model is required to judge whether there is a competitive relationship between the two focal copmanies. Considering Vicuna-7b is the backbone of our model and the better performance in the supply relation prediction task compared with closed-source LLMs, we only choose the Vicuna series model as our baseline in this task.

\autoref{tab:competitor-prediction} shows Accuracy and F1-score on a balanced set of company pairs,
comparing performance with and without SIC industry labels in the prompt. 
Despite no task-specific training, our \textbf{InterCorpRel-LLM}  
exhibit non-trivial skill in assessing competitive relationships.

Among open-source LLMs, Vicuna-7B gets a F1 of around 0.6535
and the 13B Vicuna gets a lower F1 around 0.5. Moreover, including the industry 
labels (SIC sector names) in the prompt gave the two model opposite marginal effects.
Our \textbf{InterCorpRel-LLM\_IC\_SRP}, on the other hand, generalize remarkably well to this zero-shot competitor identification challenge, with an F1-score of 0.7774 on the balanced competitor dataset. Still, steady performance improvement can be found when the SIC information is added. For the Stage I-only model \textbf{InterCorpRel-LLM\_CGM}, as expected, again underperforms on this task with a performance slightly above random.

This finding indicates that our model has internalized a robust conception of business competition, despite not being explicitly trained on a competitor identification task. It suggests that \textbf{InterCorpRel-LLM} has acquired generalizable reasoning patterns about inter-firm relationships through exposure to domain-specific data and carefully designed training objectives. Such a capability holds particular value for applications in market analysis and strategic intelligence. 

\section{Conclusion}

Our experiments highlight the importance of integrating both textual and graph-based knowledge for modeling inter-firm relationships. The traditional GNN (structure-only) and LLM (text-only) baselines struggled to achieve high recall and precision on inter-firm relationship identification tasks, particularly in out-of-distribution settings. In contrast, \textbf{InterCorpRel-LLM}, through its two-stage multi-modal training, achieved substantially higher accuracy on supply relation prediction, and this capability generalized effectively to the new task of competitor identification. Despite using only a 7B-parameter backbone, our model outperforms much larger systems—including GPT-5 and DeepSeek-v3.1—achieving an F-score of 0.8543 versus 0.2287 in supply relation prediction. Notably, we accomplish this without adjusting the backbone parameters: only a lightweight graph–language projector is trained, demonstrating the parameter efficiency and scalability of our approach. These results underscore that carefully aligning large language models with domain-specific graph knowledge yields robust, transferable understanding of corporate networks, paving the way for more intelligent and efficient business analytics tools.

\bibliographystyle{ACM-Reference-Format}
\bibliography{sample-base}

\appendix

\section{Company Description Sample}

The company descriptions following are all result of original company annual reports processed by Deepseek-V3.

\textbf{MongoDB, Inc.:} MongoDB, Inc. is a leading developer data platform company that empowers organizations to innovate through software and data. The company offers an integrated suite of database and related services designed to support modern application development across various industries and use cases. MongoDB's flagship product is its document-based database, which combines the flexibility of non-relational databases with key features of traditional relational databases, enabling developers to build and scale applications efficiently.  

The company operates primarily through two key offerings: **MongoDB Atlas**, a fully managed multi-cloud database-as-a-service (DBaaS) solution, and **MongoDB Enterprise Advanced**, a self-managed commercial database for enterprise deployments. MongoDB Atlas provides automated provisioning, monitoring, and security, allowing customers to focus on application development rather than infrastructure management. Enterprise Advanced caters to organizations requiring on-premises, hybrid, or customized cloud deployments with advanced security and management features.  

MongoDB serves a diverse customer base, including enterprises across industries such as financial services, healthcare, retail, and technology. The company follows a developer-centric approach, fostering a large global community through free offerings like **Community Server** and a **MongoDB Atlas free tier**, which help drive adoption and eventual conversion to paid subscriptions.  

With a strong focus on innovation, MongoDB continues to expand its platform with additional capabilities such as **search, time-series data handling, mobile synchronization, and analytics integrations**, reducing the need for multiple specialized database solutions. The company operates globally, with significant revenue generated outside the U.S., and maintains strategic partnerships with major cloud providers (AWS, Google Cloud, and Microsoft Azure) and system integrators to enhance market reach and customer adoption.  

MongoDB's growth strategy includes acquiring new customers, expanding within existing accounts, extending product leadership, and strengthening its developer community and partner ecosystem. The company is publicly traded on Nasdaq under the symbol **MDB**.

\textbf{Amplify Energy Corp.:} Amplify Energy Corp. is an independent oil and natural gas company engaged in the acquisition, development, and production of oil and natural gas properties across key U.S. regions, including Oklahoma, the Rockies (Bairoil), federal waters offshore Southern California (Beta), East Texas/North Louisiana, and the Eagle Ford. The company operates primarily in mature oil and gas reservoirs, focusing on both operated and non-operated working interests in producing and undeveloped leasehold acreage.  

As of December 31, 2022, Amplify Energy reported total estimated proved reserves of approximately 124.0 million barrels of oil equivalent (MMBoe), with a significant portion classified as proved developed reserves. The company's production mix includes natural gas (42\%), oil (39\%), and natural gas liquids (NGLs) (19\%). Amplify operates a substantial portion of its assets, managing properties that account for 92\% of its total proved reserves.  

Key operational segments include:  
- **Oklahoma**: Focused on wells in Alfalfa and Woods counties, contributing 28\% of proved reserves.  
- **Rockies (Bairoil)**: Primarily located in Wyoming’s Lost Soldier and Wertz fields, accounting for 23\% of reserves.  
- **Southern California (Beta)**: Offshore production platforms (Ellen, Eureka, and Elly) with a 16-inch pipeline, contributing 11\% of reserves (currently non-producing due to a pipeline incident in October 2021).  
- **East Texas/North Louisiana**: Includes fields such as Joaquin and Carthage, representing 35\% of reserves.  
- **Eagle Ford (Non-Op)**: Non-operated assets in the Eagleville fields, making up 2\% of reserves.  

Amplify Energy markets its production under month-to-month contracts with major customers, including HF Sinclair Corporation, Southwest Energy LP, and Koch Energy Services. The company relies on third-party midstream services for NGL commitments in Oklahoma.  

The company’s operations are subject to commodity price volatility, regulatory oversight, and operational risks, including the recent pipeline incident in Southern California. Amplify Energy continues to focus on reserve replacement, cost management, and strategic development to sustain production and financial performance.

\section{Experiment PROMPTS}

\textbf{Prompt for extracting key information from corporate annual reports} :

You are a professional financial analysis assistant. Based on the provided information on the company, generate an English business description that describes the main business model, the segments the company operates in and the products the company offers. The description should be written from an outsider’s perspective. Do not use other information you may have on the company. The description should not exceed 200 tokens. Just provide the description, do not add further comments. Please summarize the following important information in the company's annual report:\{text\}.

\textbf{Prompt for Company Graph Matching Task for InterCorpRel-LLM} \textit{(\textless graph\textgreater  is a placeholder that will be replaced by graph tokens generated by the Alignment Module. \{company\_names\} is the text sequence of company names to be matched.)}  : 

Given a sequence of graph tokens \textless graph\textgreater  that constitute a subgraph of an industry supply chain graph, where the first token represents the central node of the subgraph, and the remaining nodes represent the first order neighbors of the central node. Each graph token contains the content of the introduction of the company. If a company supply some product or some service to another one, a link is constructed from the former company to the other one. Here is a name list of companies:\{company\_names\}, please reorder the name list according to the order of graph tokens (i.e., complete the matching of graph tokens and company-name list).

\textbf{Prompt for Industry Classification Task for InterCorpRel-LLM} \textit{(\textless graph\textgreater  is a placeholder that will be replaced by graph tokens generated by the Alignment Module. \{company\_description\} is the focal company textual description. \{company\_name\} is the the focal company's name.)}:

Given an industry supply chain graph: \textless graph\textgreater where the 0th node is the central company, and other nodes are its one-hop or multi-hop neighbors, with the following information: Description: \{ company\_description \} Company name: \{ company\_name \} Question: Which industry category does the company belong to under the SIC classification system? Give the most likely SIC industry category of this company directly, in the form "full name of the category" .

\textbf{Prompt for Supply Relation Prediction Task for InterCorpRel-LLM} \textit{(\textless graph1/2\textgreater  is a placeholder that will be replaced by graph tokens generated by the Alignment Module. \{company1/2\_description\} is the focal company textual description. \{company1/2\_name\} is the the focal company's name.)}:

Given a sequence of graph tokens: \textless graph1\textgreater that constitute a subgraph of an industry supply chain graph, where the first token represents the central node of the subgraph, and the remaining nodes represent the first order neighbors of the central node. The information of the central node is as follow: Description: \{company1\_description\} Company name: \{company1\_name\}. and the other sequence of graph tokens: \textless graph2\textgreater , where the first token (the central node) with the following information: Description: \{company2\_description\}. Company name: \{company2\_name\}. If the link from node 1 to node 2 represent the supply chain relationship from the former company to the latter company, is there a link from node 1 to node 2? Give me a direct answer of "yes" or "no".

\textbf{Prompt for Competitor Identification Task for InterCorpRel-LLM} \textit{(\textless graph1/2\textgreater  is a placeholder that will be replaced by graph tokens generated by the Alignment Module. \{company1/2\_description\} is the focal company textual description. \{company1/2\_name\} is the the focal company's name.)}:

Given a sequence of graph tokens: \textless graph1\textgreater that constitute a subgraph of an industry supply chain graph, where the first token represents the central node of the subgraph, and the remaining nodes represent the first order neighbors of the central node. The information of the central node is as follow: Description: \{company1\_description\} Company name:  \{company1\_name\}. and the other sequence of graph tokens: \textless graph2\textgreater , where the first token (the central node) with the following information: Description:  \{company2\_description\}. Company name: \{company2\_name\} . If the link of the industry supply chain graph represents a supply chain relationship from the source company to the target company, whether the two companies represented by the central nodes of the two subgraphs are competitors of each other in business? Give me a direct answer of "yes" or "no".

\textbf{Prompt for Supply Relation Prediction Task for LLMs} \textit{(\{company1/2\_description\} is the focal company textual description. \{company1/2\_name\} is the the focal company's name. \{company1/2\_connections\} is a text description of the focal company's first-tier suppliers and first-tier customers.)}:

Given two graph nodes which are both subgraphs from a graph of an industry supply chain graph. The information of the first central node is as follow: Description: \{company1\_description\} Company name: \{company1\_name\}. Connections: \{company1\_connections\} and the second central node with the following information: Description: \{company2\_ description\}. Company name: \{company2\_name\}. Connections: \{company2\_connections\} If the link from node 1 to node 2 represent the supply chain relationship from the former company to the latter company, Is there a link from node 1 to node 2? Give me a direct answer of "yes" or "no".

\textbf{Prompt for Competitor Identification Task for LLMs} \textit{(\{company1/2\_description\} is the focal company textual description. \{company1/2\_name\} is the the focal company's name. \{company1/2\_connections\} is a text description of the focal company's first-tier suppliers and first-tier customers.)}:

Given two sequences of graph nodes which are both subgraphs from a graph of an industry supply chain. The information of the first central node in the first subgraph is as follow: Description: \{company1\_description\} Company name: \{company1\_name\}. Connections: \{company1\_connections\} and the second central node in the other subgraph with the following information: Description: \{company2\_description\} Company name: \{company2\_name\}. Connections: \{company2\_connections\} If the link of the industry supply chain graph represents a supply chain relationship from the source company to the target company, whether the two companies represented by the core nodes of the two subgraphs are competitors of each other in business? Give me a direct answer of "yes" or "no".

\clearpage

\section{Company Data Sample}

\begin{table*}[!b]
\centering
\scriptsize
\renewcommand{\arraystretch}{1.1}
\caption{Company Data Sample}
\label{tab:comp-rel-full}
\begin{tabularx}{\textwidth}{c l c l l X}
\toprule
\textbf{Year} & \textbf{Name} & \textbf{Country} & \textbf{Province} & \textbf{CIK} & \textbf{SIC\_Label} \\
\midrule
2023 & A. O. Smith Corp. & US & Wisconsin & 91142 & Household Appliances \\
2023 & Aziyo Biologics, Inc. & US & Maryland & 1708527 & Biological Products, except Diagnostic Substances \\
2023 & BayCom Corp. & US & California & 1730984 & Commercial Banks \\
2023 & Co-Diagnostics, Inc. & US & Utah & 1692415 & In Vitro and In Vivo Diagnostic Substances \\
2023 & DaVita, Inc. & US & Colorado & 927066 & Miscellaneous Health and Allied Services, Not Elsewhere Classified \\
2023 & Ikena Oncology, Inc. & US & Massachusetts & 1835579 & Pharmaceutical Preparations \\
2023 & Inozyme Pharma, Inc. & US & Massachusetts & 1693011 & Biological Products, except Diagnostic Substances \\
2023 & Inspired Entertainment, Inc. & US & New York & 1615063 & Motion Picture and Video Tape Production \\
2023 & Invitation Homes, Inc. & US & Texas & 1687229 & Real Estate Investment Trusts \\
2023 & Kadant Inc. & US & Massachusetts & 886346 & Special Industry Machinery, except Metalworking \\
2023 & Kaspien Holdings, Inc. & US & Washington & 795212 & Catalog and Mail-Order Houses \\
2023 & Kimbell Royalty Partners LP & US & Texas & 1657788 & Crude Petroleum and Natural Gas \\
2023 & Lantern Pharma, Inc. & US & Texas & 1763950 & Biological Products, except Diagnostic Substances \\
2023 & Microbot Medical, Inc. & US & Massachusetts & 883975 & Electromedical and Electrotherapeutic Apparatus \\
2023 & Southwest Gas Holdings, Inc. & US & Nevada & 1692115 & Natural Gas Transmission and Distribution \\
2023 & Stoke Therapeutics, Inc. & US & Massachusetts & 1623526 & Biological Products, except Diagnostic Substances \\
2023 & TESSCO Technologies, Inc. & US & Maryland & 927355 & Electronic Parts and Equipment, Not Elsewhere Classified \\
2023 & TETRA Technologies, Inc. & US & Texas & 844965 & Oil and Gas Field Services, Not Elsewhere Classified \\
2023 & TSR, Inc. & US & New York & 98338 & Computer Programming Services \\
2023 & TTEC Holdings, Inc. & US & Texas & 1013880 & Business Services, Not Elsewhere Classified \\
2023 & Tejon Ranch Co. & US & California & 96869 & Crops \\
2023 & Teleflex, Inc. & US & Pennsylvania & 96943 & Surgical and Medical Instruments and Apparatus \\
2023 & Telephone \& Data Systems, Inc. & US & Illinois & 1051512 & Radiotelephone Communications \\
2023 & Tenet Healthcare Corp. & US & Texas & 70318 & General Medical and Surgical Hospitals \\
2023 & Tennant Co. & US & Minnesota & 97134 & Refrigeration and Service Industry Machinery \\
2023 & Tetra Tech, Inc. & US & California & 831641 & Engineering Services \\
2023 & Texas Instruments Incorporated & US & Texas & 97476 & Semiconductors and Related Devices \\
2023 & Textron, Inc. & US & Rhode Island & 217346 & Aircraft \\
2023 & The Charles Schwab Corp. & US & Texas & 316709 & Investment Advice \\
2023 & The TJX Cos., Inc. & US & Massachusetts & 109198 & Family Clothing Stores \\
2023 & The Timken Co. & US & Ohio & 98362 & Ball and Roller Bearings \\
2023 & Thermo Fisher Scientific, Inc. & US & Massachusetts & 97745 & Laboratory Analytical Instruments \\
2023 & ThermoGenesis Holdings, Inc. & US & California & 811212 & Laboratory Apparatus and Furniture \\
2023 & Thor Industries, Inc. & US & Indiana & 730263 & Miscellaneous Transportation Equipment \\
2023 & Tidewater, Inc. & US & Texas & 98222 & Water Transportation \\
2023 & Titan International, Inc. & US & Illinois & 899751 & Farm Machinery and Equipment \\
2023 & Tofutti Brands, Inc. & US & New Jersey & 730349 & Ice Cream and Frozen Desserts \\
2023 & Tractor Supply Co. & US & Tennessee & 916365 & Building Materials, Hardware, Garden Supply, and Mobile Home Dealers \\
2023 & Trans-Lux Corp. & US & New York & 99106 & Miscellaneous Manufacturing Industries \\
2023 & TransAct Technologies, Inc. & US & Connecticut & 1017303 & Computer Peripheral Equipment, Not Elsewhere Classified \\
2023 & Transcat, Inc. & US & New York & 99302 & Testing Laboratories \\
2023 & Transcontinental Realty Investors, Inc. & US & Texas & 733590 & Operators of Apartment Buildings \\
2023 & Tredegar Corp. & US & Virginia & 850429 & Rolling, Drawing, and Extruding of Nonferrous \\
2023 & Trimble, Inc. & US & Colorado & 864749 & Measuring and Controlling Devices, Not Elsewhere Classified \\
2023 & Trinity Industries, Inc. & US & Texas & 99780 & Railroad Equipment \\
\bottomrule
\end{tabularx}
\end{table*}

\end{document}